\documentclass[times, review, 10pt]{elsarticle}

\usepackage{setspace}
\usepackage[numbers]{natbib}
\usepackage{amsmath,amsfonts}
\usepackage{color}
\usepackage{xcolor}
\usepackage{booktabs}
\usepackage{multirow}
\usepackage{multirow}
\usepackage{pifont}
\usepackage{colortbl, booktabs}
\usepackage{array}
\usepackage{graphicx}
\usepackage{subcaption}
\usepackage{caption}
\usepackage{float}
\usepackage[table]{xcolor}
\usepackage[normalem]{ulem}
\usepackage{amssymb}
\usepackage{amsmath}

\usepackage{booktabs}
\usepackage{multirow}
\usepackage{amssymb}
\usepackage{array}
\usepackage{paralist}       
\usepackage{amsmath} 
\usepackage{url}
\usepackage{algorithm}
\usepackage{algpseudocode} 

\usepackage{stfloats}
\usepackage{subcaption}

\journal{Elsevier}

\usepackage{xspace}
\newcommand{\alias}{REMAC\xspace}

\begin{document}
\definecolor{gray}{rgb}{0.5, 0.5, 0.5}  
\begin{frontmatter}



\title{REMAC: Self-Reflective and Self-Evolving Multi-Agent Collaboration for Long-Horizon Robot Manipulation}






\author[1]{Puzhen Yuan}
\ead{ypz21@mails.tsinghua.edu.cn}

\author[2]{Angyuan Ma}
\ead{maay21@mails.tsinghua.edu.cn}

\author[3]{Yunchao Yao}
\ead{yunchaoy@andrew.cmu.edu}

\author[4]{Huaxiu Yao}
\ead{huaxiu@cs.unc.edu}

\author[5]{Masayoshi Tomizuka}
\ead{tomizuka@berkeley.edu}

\author[4,5]{Mingyu Ding\corref{mycorrespondingauthor}}
\cortext[mycorrespondingauthor]{Corresponding author}
\ead{md@cs.unc.edu}

\address[1]{Institute of Interdisciplinary Information Sciences at Tsinghua University, Beijing, 100084, China}
\address[2]{Department of Automation at Tsinghua University, Beijing, 100084, China}
\address[3]{Robotics Institute at Carnegie Mellon University, Pittsburgh, 15213, America}
\address[4]{Department of Computer Science at UNC-Chapel Hill, Chapel Hill, 27599, America}
\address[5]{Department of Mechanical Engineering at UC Berkeley, Berkeley, 94720, America}


\begin{abstract}
Vision-language models (VLMs) have demonstrated remarkable capabilities in robotic planning, particularly for long-horizon tasks that require a holistic understanding of the environment for task decomposition.
Existing methods typically rely on prior environmental knowledge or carefully designed task-specific prompts, making them struggle with dynamic scene changes or unexpected task conditions. 
Such challenges underscore two critical issues: adaptability and efficiency.
To address them, in this work, we propose an adaptive multi-agent planning framework, termed \alias, that enables efficient, scene-agnostic multi-robot long-horizon task planning and execution through continuous reflection and self-evolvement.
\alias incorporates two key modules: a self-reflection module performing pre-condition and post-condition checks in the loop to evaluate progress and refine plans, and a self-evolvement module dynamically adapting plans based on scene-specific reasoning.
It offers appealing benefits: 
1) Robots can keep reflecting on potential planning errors and adapting the plan based on task-specific insights.
2) After iterations, a robot can call another one to coordinate tasks in parallel, maximizing the task execution efficiency.
To validate \alias's effectiveness, we build a multi-agent environment for long-horizon robot manipulation and navigation based on RoboCasa, featuring 4 task categories with 27 task styles and 50+ different objects.
Based on it, we further benchmark state-of-the-art reasoning models, including DeepSeek-R1, o3, QwQ, Qwen3, and Grok3.
Extensive experiments demonstrate \alias's superiority by boosting average success rates by 40\% and execution efficiency by 52.7\% over the single robot baseline without any task-specific prompting or fine-tuning in initial planning.
Code, demos, and appendix are available at the website https://reamac-repo.github.io/remac-project-page/.
\end{abstract}

\begin{keyword}
self-reflection \sep self-evolvement \sep multi-agent collaboration \sep vision-language model \sep robotic manipulation 
\end{keyword}

\end{frontmatter}

\section{Introduction}

In recent years, Vision-Language Models (VLMs) have seen significant application in robot control tasks~\cite{li2024vision,huang2022language}.
VLMs combine visual information with language descriptions, enabling robots to understand and execute complex tasks, thereby demonstrating strong planning capabilities. 
In household environments, robots often face a series of complex, long-horizon tasks, such as preparing food or organizing cabinets~\cite{triantafyllidis2023hybrid}. These tasks typically involve multiple stages and require the robot to have a comprehensive understanding of the environment. Additionally, tasks such as cooking may require collaboration between multiple robots to ensure efficiency and quality. In such situations, how to guide multiple robots to correctly plan and execute tasks using VLMs remains a key challenge.

\begin{figure}
  \centering
  \includegraphics[width=1.0\linewidth]{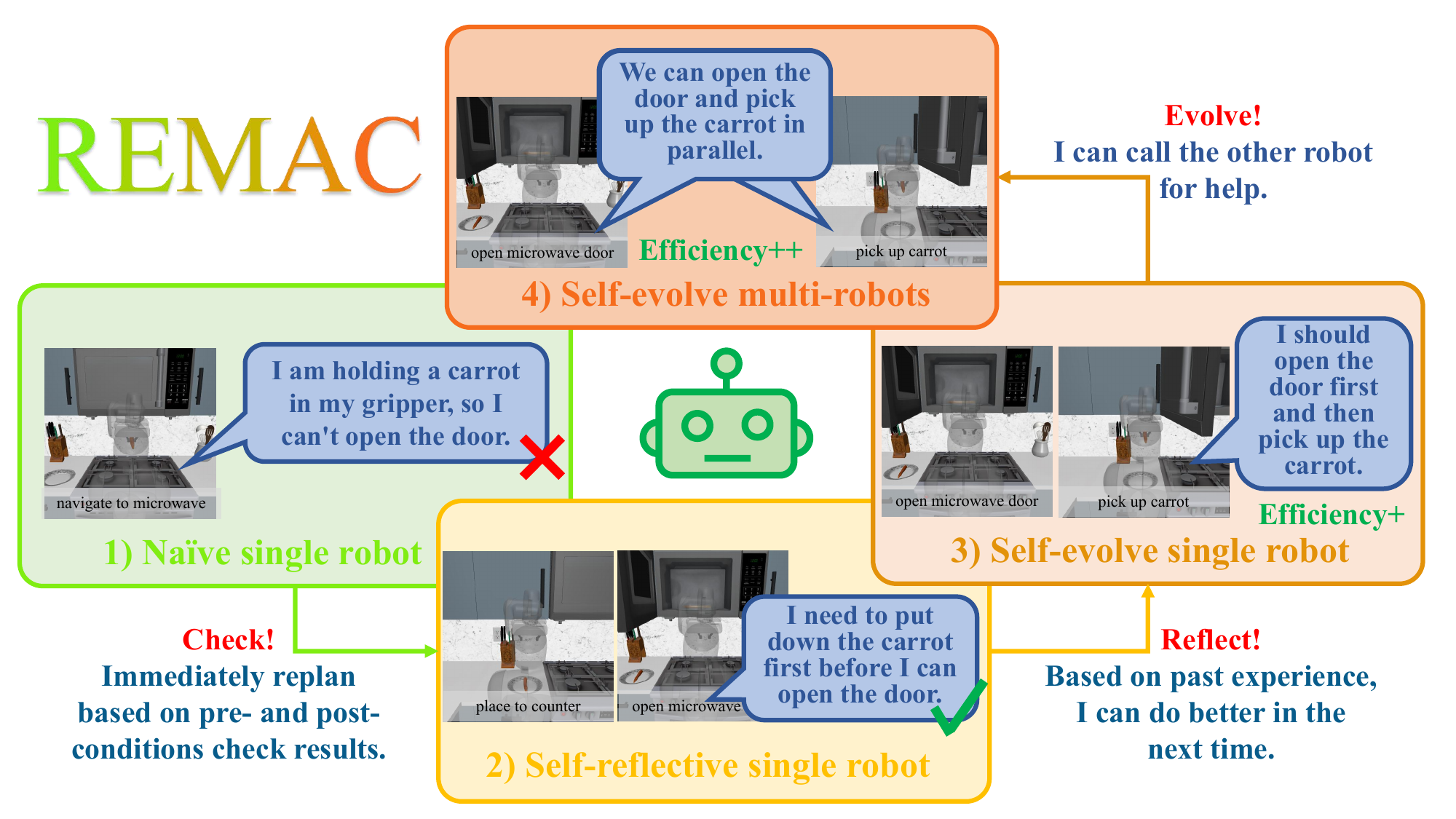}
  \caption{\alias resolves planning errors in long-horizon, multi-stage tasks by employing reflection given condition check results and evolvement based on reflection, thereby enhancing planning efficiency iteratively.  \textbf{1)} Task fails because the robot attempts to open the microwave door while holding a carrot. \textbf{2)} Task succeeds after the robot, through a condition check process, recognizes the necessity of putting down the carrot. \textbf{3)} Efficiency is enhanced when the robot correctly sequences its actions by opening the microwave door first and then picking the carrot up. \textbf{4)} Efficiency is further improved when the robot delegates the task of opening the door to another robot.
}
  \label{fig:teaser}
\end{figure}

Existing long-horizon task planning methods often take robot observation images and complex prompts containing scene information as input and output multi-stage task plans~\cite{wake2024gpt,wu2024mldt}.
However, these models require comprehensive knowledge of both the environment and the task, typically provided through complex, scene-specific prompts. Without such detailed input, VLMs may struggle to infer critical spatial and logical constraints, leading to infeasible or inefficient task plans \cite{zhuo2022vilpact}. Furthermore, relying on VLMs to generate a single, fixed plan for an entire task is often inefficient. Any change in the environment or failure in sub-task execution can lead to complete re-planning, making the approach fragile and inflexible for dynamic scenarios.
On the other hand, for tasks requiring bimanual coordination, current approaches usually involve splitting complex tasks and assigning them to multiple robots for synchronous execution~\cite{liu2024towards,talebirad2023multi}.
These methods can only tackle short-horizon tasks and lack mechanisms to dynamically refine initial plans, limiting their effectiveness in long-horizon tasks~\cite{zhang2024lamma}.
Therefore, guiding multiple robots to complete long-horizon, multi-stage tasks accurately remains a significant challenge.

In this paper, we propose a novel framework for long-horizon, multi-robot task planning called \alias, which enables the agent to continuously interact with the environment and improve planning through self-reflection and self-evolvement, as shown in Figure~\ref{fig:teaser}. 
\alias is a zero-shot framework that generates efficient and feasible multi-robot collaborative task plans based on concise task descriptions provided by humans across diverse scenarios. 
The final task plans are generated through iterative progress in reflection and evolvement. 
Specifically, based on scene information and task goal, a VLM will generate the initial plan for robots to execute.
When executing each sub-task, the VLM checks pre-conditions to reflect on any potential planning errors based on the scene, and checks post-conditions to guarantee successful execution.
This reflection mechanism effectively captures key scene information, enabling the VLM to generate more reasonable plans and preventing it from overlooking key constraints when processing large amounts of information.

Figure~\ref{fig:teaser} illustrates the proposed \alias, which allows the VLM to correct task planning errors based on scene information, guiding multiple robots to collaboratively complete task plans that align with the scene’s constraints. 
To validate \alias's effectiveness, we build a multi-agent benchmark for long-horizon robot manipulation and navigation based on RoboCasa, with 4 task categories, 27 task styles and 50+ different objects. 
Extensive experiments on this demonstrate the effectiveness of our framework.
We also benchmark latest reasoning models including DeepSeek-R1, o3, QwQ, Qwen3, and Grok3 on their reflective ability.

In summary, our contributions are as follows.
\begin{itemize}
    \item We propose a multi-robot collaborative planning framework with self-reflection and self-evolvement capabilities for long-horizon and multi-stage tasks.
    \item We introduce a self-evolvement mechanism leveraging pre- and post-condition checking modules for both in-the-loop task re-planning and future task optimization. %
    \item We build a multi-robot simulation environment based on RoboCasa that provides physical controllers, together with a comprehensive long-horizon benchmark evaluating state-of-the-art reasoning models.
    \item Extensive experiments demonstrate the effectiveness of our framework, \emph{e.g.}, improve the average success rate by 40.0\% and efficiency by 52.7\% compared to a single robot baseline.
\end{itemize}

\begin{figure}
  \centering
  \includegraphics[width=0.99\linewidth]{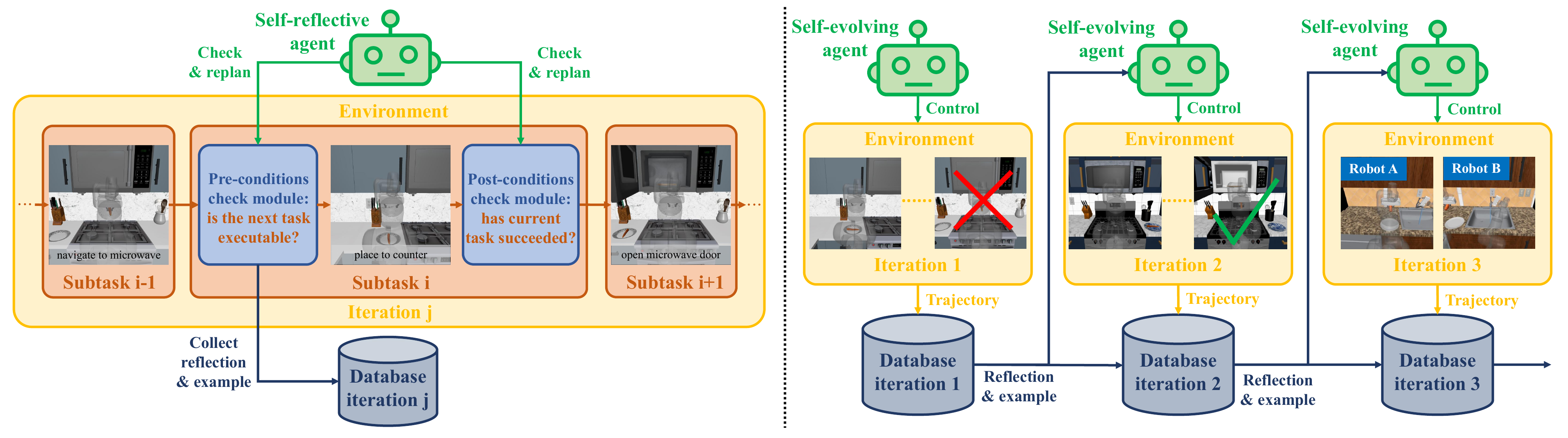}
  \caption{\textbf{Left: Self-Reflection.} Before the execution of subtask $i$, the VLM verifies the pre-conditions to determine whether the plan for subtask $i$ is executable given the observation after completing subtask $i-1$. If not, this indicates an error in the initial planning, and the system engages in a reflection process to identify the cause of this error, which is subsequently stored in the reflection database. Following the execution of subtask $i$, the VLM verifies the post-conditions to assess whether the subtask was successfully executed, given the observation after executing the current task. \textbf{Right: Self-Evolvement.}Upon sequential completion of all subtasks, the reflection database—containing accumulated pre-condition-check analysis and last-iteration plan serves as the foundation for generating initial plans for subsequent iterations. This knowledge-augmented process iteratively refines planning logic, yielding an optimized initial plan with feasibility and efficiency, and is used in the prompt of multi-agent collaboration to further improve efficiency.} 
  \label{figure:framework}
\end{figure}

\section{Related Work}

\subsection{VLM-Based Robot Task Planning}

Vision-Language Models (VLMs) transform inputs from natural language and images into formal languages that model-based planners can interpret~\cite{yang2024guiding,tang2026rethinking}
Recent studies have demonstrated the potential of language models for robot control. SayCan~\cite{ahn2022saycan} introduced a framework that grounds language model reasoning into robot affordances by combining linguistic knowledge with executable skills. Grounded Decoding~\cite{huang2023grounded} further incorporated grounded models into language generation, enabling embodied agents to generate actions consistent with environmental constraints. 
VoxPoser~\cite{huang2023voxposer} extended this direction by constructing composable 3D value maps from language instructions, allowing language models to reason about spatial affordances for manipulation.
RoboAgent~\cite{xu2026roboagent} proposes a capability-driven planning framework that decomposes embodied tasks into a sequence of reusable vision-language capabilities, enabling more transparent and controllable reasoning.
However, a significant challenge emerges when VLMs fail to account for contextual details provided by scene or image information, resulting in outputs that are impractical or unfeasible in real-world settings.

To improve robustness against planning failures, several works have explored reflection and replanning mechanisms. RePlan~\cite{skreta2024replan} proposed a perception-language based replanning framework that allows robots to revise task plans according to updated environmental observations. 
REFLECT~\cite{liu2023reflect} introduced a failure explanation mechanism that summarizes robot experiences and uses these explanations to correct future decisions. 
LLM³~\cite{wang2024llmˆ} further incorporated motion failure reasoning into language-model-based task and motion planning, enabling robots to analyze execution failures at both task and motion levels. 
Reflective Planning~\cite{feng2025reflective} introduces an iterative reflection framework that improves VLM-based long-horizon manipulation through imagined future states and self-evaluation.
However, these methods face substantial difficulties when applied to tasks that span extended durations and comprise numerous subtasks, especially under coarse-grained and reasoning-difficult instructions in daily life (e.g. ``heat the food'' or ``defrost the meat'').

Beyond failure recovery, recent studies have investigated the use of large language models for long-horizon robotic planning. Inner Monologue~\cite{pmlr-v205-huang23c} proposed a framework that integrates language models with environmental feedback, enabling robots to iteratively reason about task execution. ISR-LLM~\cite{zhou2024isr} introduced an iterative self-refinement mechanism for sequential task planning, where the language model improves generated plans through repeated refinement.
Mind Palace~\cite{ginting2025enter} introduces structured episodic memory to support long-term embodied reasoning by retrieving past experiences for future decision making.
Interactive Task Planning with Language Models~\cite{li2025interactive} explored interactive planning strategies that combine human instructions and environmental feedback to solve complex manipulation tasks, but still suffer from vague instructions.
In response to the limitation above, our proposed method, denoted as \alias, overcomes the inability of VLMs to reflect on and replan for high-level language instructions under ambiguous instructions. By doing so, it effectively extends the reasoning capabilities of VLMs to accommodate complex tasks characterized by prolonged temporal horizons.

\subsection{Multi-Robot Collaboration}

Task planning in multi-robot systems is crucial for improving task execution efficiency \cite{chakraa2023optimization}. 
More recently, Li et al. \cite{li2026large} systematically reviewed the emerging use of LLMs in multi-robot systems, covering communication, task allocation, task planning, and human-robot interaction.
Typically, task planning for multi-robot systems involves two stages: task decomposition and task allocation. In the task decomposition stage, a convenient and cognitively aligned method is to describe the overall task using high-level natural language.
TACO~\cite{shiarlis2018taco} learned task decomposition through temporal alignment between tasks and control trajectories, providing a data-driven approach for discovering task structures. 
Motes et al.~\cite{motes2020multi} studied multi-robot task and motion planning with explicit subtask dependencies, enabling coordinated execution among robots. 
SMART-LLM~\cite{kannan2024smart} introduced an LLM-driven framework that decomposes high-level instructions into robot-executable subtasks and assigns them according to robot capabilities. 
RoboMatrix~\cite{mao2024robomatrix} proposed a skill-centric hierarchical framework that scales robot task planning by organizing reusable skills in open-world environments.
Murata et al.~\cite{murata2026multi} further leveraged LLMs to integrate distributed on-site knowledge across robots for collaborative task planning in multi-object retrieval scenarios.
 
However, existing approaches generally assume that generated subtasks are feasible before execution. Text2Motion~\cite{lin2023text2motion} investigated the translation from natural language instructions to feasible motion plans, highlighting the challenge of satisfying physical constraints during planning. Rizk et al.~\cite{rizk2019cooperative} provided a survey of cooperative heterogeneous multi-robot systems and discussed the challenges of coordination, scalability, and robustness in practical deployments.
Our work addresses this limitation by introducing a reflection process, originally developed for single-robot systems, into multi-robot systems. This approach enhances the efficiency of the entire robot system using a zero-shot methodology.

\section{Problem Statement}

Given a high-level language instruction $I$ and an environment $E$, our objective is to understand the layout of $E$ and the objects it contains and to decompose $I$ into multiple subtasks that can be assigned to multiple robots for efficient execution. Notably, to ensure generality, the language instruction $I$ is designed to be independent of any specific scene, which poses a significant challenge for robot movement and operation tasks. For example, consider the instruction ``heat the vegetables.'' Directly querying LLM for task decomposition might lead to arbitrary and irrelevant task plans, such as "open microwave" being a subtask in an environment with only a pan and a stove. 

In this work, we consider a set of robots $R = \{R_i\}_{i=1,\dots,n}$ ($n = 1$ or $2$ in our case) operating within the scene $E$, with the goal of decomposing a high-level language instruction $I$ into $K$ subtasks $T = \{T_i\}_{i=1,\dots,K}$ that can be completed using the items available in $E$. Our approach focuses on improving the understanding of the scene $E$ to provide sufficient information for accurate task decomposition and to prevent failures due to planning errors, while also addressing efficient allocation of subtasks with the same temporal priorities to multiple robots for execution in parallel, particularly since some subtasks can be performed concurrently.

\section{Method}

In this section, we elaborate on the specifics of our proposed framework which comprises two interconnected stages. Given scene information and task goal, a large language model decomposes the high-level task description into subtasks that can be executed by low-level policies. For each subtask, assessments are also conducted to evaluate their feasibility and the overall progress toward completion, with pertinent reflections being systematically recorded in a buffer. Finally, leveraging the accumulated reflections, the initial plan is iteratively refined and advanced through multiple cycles of optimization. 

\subsection{Task Planning and Check Mechanism}

In this stage, the items listed in Scene $E$ serve as a reference for a large language model with reasoning capabilities to decompose tasks based on language instructions. For example, for the task of "heating carrots", since the items listed in Scene $E$ include a microwave, the task is naturally broken down into subtasks such as placing the carrots in the microwave and starting it. 
To further improve efficiency, the LLM also analyzes the temporal dependencies of subtasks and assigns them to multiple robots whenever parallel execution is possible.
For instance, with the task instruction of ``preparing a breakfast with coffee and vegetables", one robot would handle the coffee while another heats the vegetables.

To tackle the issue of VLM exhibiting insufficient attention to logical and spatial constraints in task planning, inspired by \cite{zhou2023generalizable}, we introduced a more effective mechanism: \textbf{1)} pre-conditions check to assess the feasibility of the plan and \textbf{2)} post-condition checks to evaluate the successful execution of subtasks, as shown in Figure~\ref{figure:framework} left.

\textbf{Pre-conditions Check.} Long-horizon tasks may fail due to planning errors at any step. We implement pre-condition checks to verify subtask feasibility. A VLM receives pre-execution scene images and subtask descriptions to assess whether current conditions meet planning requirements. If met, the robot executes the subtask; otherwise, the VLM identifies failure reasons for replanning the upcoming subtask sequence and re-evaluating pre-conditions. This reflection is stored in a buffer for iterative improvements. Please visit our website for more details and examples.

\textbf{Post-conditions Check.} Since subtask execution depends on low-level policy success rates, we implement post-condition checks for each subtask. A VLM receives post-execution scene images and subtask descriptions to evaluate whether success criteria are met. If satisfied, the robot proceeds to the next subtask; otherwise, it retries. Since unsuccessful policy execution may still alter scene conditions, invalidating prior pre-condition checks, we recheck pre-conditions as needed. Post-condition failures are attributed to policy execution errors, not planning issues, and are not used for reflection.

\subsection{Iterative Self-Evolving Framework}

Long-horizon tasks may achieve success after undergoing pre-conditions checks, reflection, post-conditions checks, and retries, but there may still be redundant steps in the execution; alternatively, the planning may remain a failure. Nevertheless, the reflections generated throughout the process are stored in a long-term memory database to guide the planning in the subsequent iteration.

Similar to the initial plan, we input the items information from Scene $E$, the language instruction $I$, and the reflection $R$ together into an LLM with reasoning abilities. The insights gained from reflecting on the reasons for failures in the previous iteration enable the LLM to generate a more reasonable plan, thereby ensuring higher success rates and efficiency. We employ multiple iterations to allow the planning to evolve through reflection, as shown in Figure~\ref{figure:framework} right.

During its self-evolving iterations, a single robot accumulates planning hints and references, which can be helpful for top-level planning of multiple robots. After the task can be successfully and efficiently executed by a single robot, we prompt the planning LLM to call the other robot to assist, further improving efficiency without sacrificing task success rate. Specifically, we prompt the main agent to extract the main-object-centric task plan and then summarize the previously recorded pre-condition sequences. We then prompt another agent to handle these pre-conditions, ensuring they are completed before the main robot, thus forming a new task plan to help the main robot improve its efficiency.

\begin{figure}
  \centering
  \includegraphics[width=1.0\linewidth]{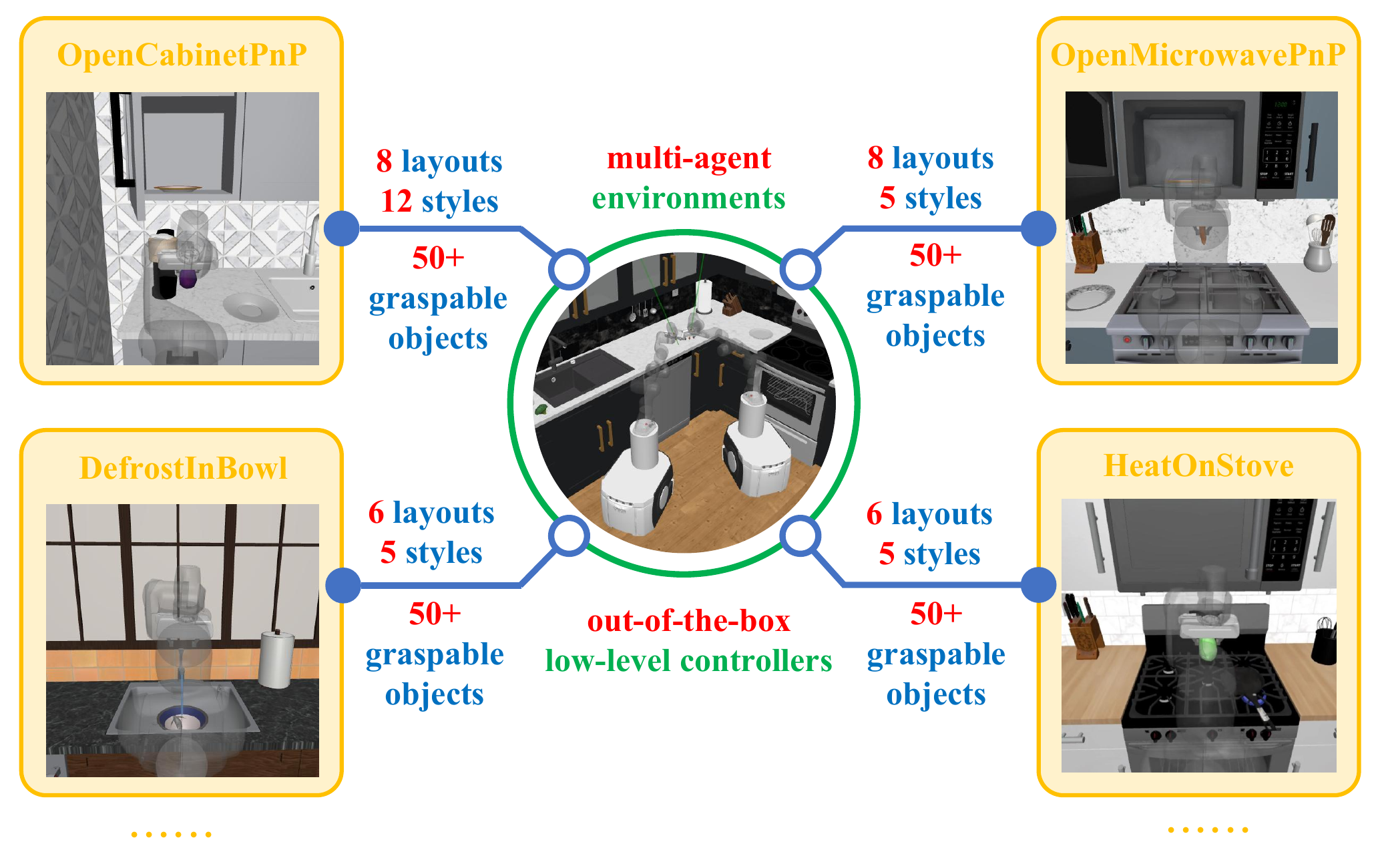}
  \caption{We constructed four distinct tasks—OpenCabinetPnP, OpenMicrowavePnP, DefrostInBowl, and HeatOnStove—within the RoboCasa environment to evaluate long-horizon multi-robot collaborative planning. Each task incorporates 6–8 spatial layouts, 5–12 dynamically configured environmental styles, and over 50 graspable objects to simulate real-world complexity. At the beginning of each trial, tasks are initialized using randomized combinations of layouts, styles, and object placements.
}
  \label{fig:benchmark}
\end{figure}

\begin{figure}
  \centering  
  \includegraphics[scale=1.0,width=1.0\textwidth]{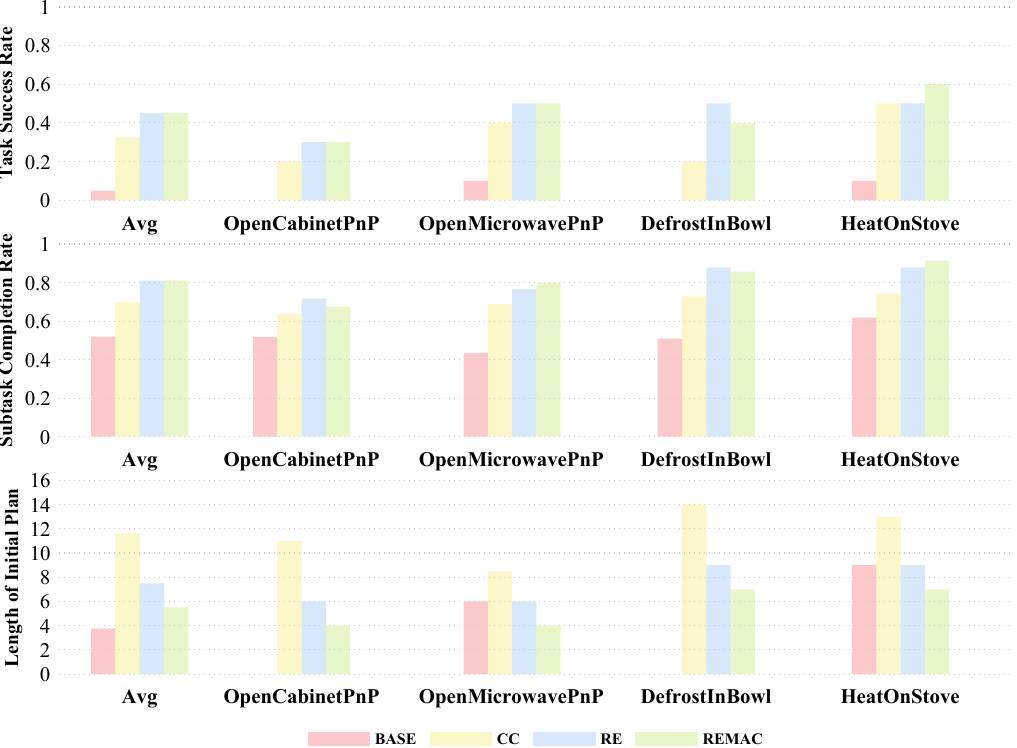} 
  \caption{Our experimental results indicate that: (1) condition checking and reflective evolvement effectively enhance both the Task Success Rate and the Subtask Completion Rate; (2) compared to single-robot systems, multi-robot systems demonstrate a reduced Length of Initial Plan and greater efficiency. All tasks were subjected to rigorous validation through 10 randomized initializations across four distinct experimental settings. The bars indicate the average value of the coresponding metrics.}
  \label{fig:comparison}
\end{figure}

\section{Experiment}

We have designed a series of experiments to address the following questions:

\begin{enumerate} 
\item How does our framework compare to the current best top-level planning methods for long-horizon and reasoning-difficult tasks?
\item To what extent do checking mechanisms grounded in pre- and post-conditions, coupled with an iterative evolvement mechanism informed by reflection, enhance the success rate over extended temporal horizons?
\item Relative to single-robot task planning, to what degree does reflection-guided multi-robot task replanning augment execution efficiency?
\end{enumerate}

\subsection{Baselines and Ablations}

We have established the following 4 experimental settings for ablation analysis:

\begin{enumerate} 
\item \textbf{single-robot planning (BASE):} A single-robot system without pre- and post-condition checking or the reflection-based iterative evolvement mechanism. 
\item \textbf{+ condition checking (CC):} A single-robot system incorporating pre- and post-condition checking.
\item \textbf{+ reflective evolvement (RE):} A single-robot system with both pre- and post-condition checking and the reflection-based iterative evolvement mechanism.
\item \textbf{+ multi-agent collaboration (\alias):} A multi-robot system integrating both pre- and post-condition checking and the reflection-based iterative evolvement mechanism.
\end{enumerate}

Considering the relevance to our reflective evolvement framework, we adopt the REFLECT~\cite{liu2023reflect} method as one of the suitable comparison baselines. REFLECT is a framework that queries LLM to replan after failure reasoning based on a hierarchical summary of the robot's past experiences. We compare our methods with REFLECT after failure correction.

We also compare with CAPE~\cite{raman2024cape}, which is a representative method for correcting action from pre-condition errors, similar to our condition checking (CC) settings. However, it does not involve self-evolvement or calling other robots for help.

Moreover, we also compare with several LLMs like gpt-4o~\cite{hurst2024gpt4o}, DeepSeek-R1~\cite{guo2025deepseek}, o3~\cite{openai2025o3}. Since these method are not equipped with low-level controllers, we use them for generating initial plans in our pipeline.

\subsection{Task Suite}

Our experiments utilize the RoboCasa framework~\cite{nasiriany2024robocasa}, which is based on the RoboSuite platform~\cite{zhu2021robosuite}. This simulation environment contains a wide variety of scenes and 3D models generated with tools such as Midjourney and Luma.AI, creating highly diverse and rich testbeds.

We designed four reasoning-difficult tasks to evaluate the effectiveness of our method in kitchen scenarios, validated using the RoboCasa large-scale simulation framework, as in Figure~\ref{fig:benchmark}.
\begin{list}{•}{\leftmargin=0.5em}
\item \textbf{OpenCabinetPnP:} To place items in the cabinet, the robot must first open its closed doors.
\item \textbf{OpenMicrowavePnP:} To heat vegetables in the microwave, the robot must first open its closed door.
\item \textbf{DefrostInBowl:} To defrost frozen meat/fish in the sink, the robot must know to use a bowl to contain them.
\item \textbf{HeatOnStove:} To heat vegetables on the stove, the robot must recognize the need for a pan.
\end{list}

\begin{table}
\centering
\setlength{\tabcolsep}{9pt}
\caption{Quantitative comparisons of \alias on four different tasks with three base settings: BASE -- single robot planning, CC -- condition checking, RE -- reflective evolvement, REMAC -- multi-agent collaboration, TSR -- task success rate, SCR -- subtask completion rate, IPL -- initial plan length.}
\label{tab:main}
\begin{tabular*}{1.0\textwidth}{@{\extracolsep{\fill}}c c c c c c}
\toprule
\textbf{Task}                      & \textbf{Setting}      & \textbf{TSR$\uparrow$} & \textbf{SCR$\uparrow$} & \textbf{Time (s)$\downarrow$} & \textbf{IPL$\downarrow$} \\ \midrule
\multirow{4}{*}{OpenCabinetPnP} & BASE &     0.00\%       &      51.67\%                &  NaN   &   NaN                \\
                       & CC &    20.00\%        &       63.64\%               &   17.0   &          11              \\
                       & RE &    \textbf{30.00\%}           &       \textbf{71.67\%}              &   9.5   &          6              \\
                       & REMAC &    \textbf{30.00\%}           &     67.50\%                 &  \textbf{7.3}    &          \textbf{4}             \\ \midrule
\multirow{4}{*}{OpenMicrowavePnP} & BASE &    10.00\%          &     43.33\%                 &   9.0   &  6                      \\
                       & CC & 40.00\%             &       68.79\%               &   14.0   &     8.5                   \\
                       & RE &  \textbf{50.00\%}            &          76.67\%            &   8.3   &               6         \\
                       &  REMAC &   \textbf{50.00\%}           &               \textbf{80.00\%}       &   \textbf{6.7}   &                 \textbf{4}       \\ \midrule
\multirow{4}{*}{DefrostInBowl} & BASE &        0.00\%      &      50.95\%                &   NaN   &           NaN             \\
                       & CC &     20.00\%         &       72.86\%               &   19.5   &         14               \\
                       & RE &     \textbf{50.00\%}         &      \textbf{87.78\%}                &   11.5   &          9              \\
                       & REMAC &    40.00\%          &       85.71\%               &   \textbf{9.7}   &       \textbf{7}                 \\ \midrule
\multirow{4}{*}{HeatOnStove} & BASE &         10.00\%     &   61.78\%                   &   15.7   &           9             \\
                       & CC &      50.00\%        &      74.29\%                &   27.5   &   13                     \\
                       & RE &     50.00\%         &         87.78\%             &   15.5   & 9                       \\
                       & REMAC &   \textbf{60.00\%}           &     \textbf{91.43\%}                 &   \textbf{12.0}   &         \textbf{7}               \\ \bottomrule
\end{tabular*}
\end{table}

To ensure our framework is generalizable, we use scene-agnostic high-level instructions. This requires the robot to first locate necessary items, after which the LLM decomposes the task into subtasks for autonomous execution. To handle complex kitchen tasks, we pre-trained language-conditioned low-level policies for common actions like pick-and-place and operating appliances (cabinets, drawers, stoves, etc.).

\subsection{Performance Metrics}

For each task, we conducted tests under four distinct conditions, each involving 10 random initializations. We use gpt-4o~\cite{hurst2024gpt4o} as the VLM to perform pre- and post-conditions checks and plan adjustments, and DeepSeek-R1~\cite{guo2025deepseek} as the reasoning-capable LLM only to reflect and generate the initial plan due to cost and latency considerations. The following three performance metrics were measured:
\begin{list}{•}{\leftmargin=0.5em}

\item Task Success Rate (TSR): The ratio of the number of trials in which all subtasks were successfully executed and the objective specified in the high-level language instruction was achieved to the total number of trials conducted.

\item Subtask Completion Rate (SCR): The ratio of the number of subtasks successfully executed by the robot to the total number of subtasks in the initial plan.

\item Initial Plan Length (IPL): Total subtasks in initial planning. Parallel subtasks by two robots count as one. Only successfully executed plans are calculated.

\end{list}

\subsection{Main Results}

For challenging long-horizon tasks, REMAC iteratively identifies causes of infeasible subtasks and reduces errors through self-evolvement. The final plan increases task success rates from near-zero baseline to 20-60\% (Figure~\ref{fig:comparison}, Table~\ref{tab:main}). The self-reflection and self-evolvement mechanisms significantly improve subtask completion ratios and overall task performance. Additionally, self-evolvement enhances planning efficiency by reducing initial plan length by 35-62\% through pruning redundant subtasks.
\begin{figure}
  \centering
  \includegraphics[width=1.0\linewidth]{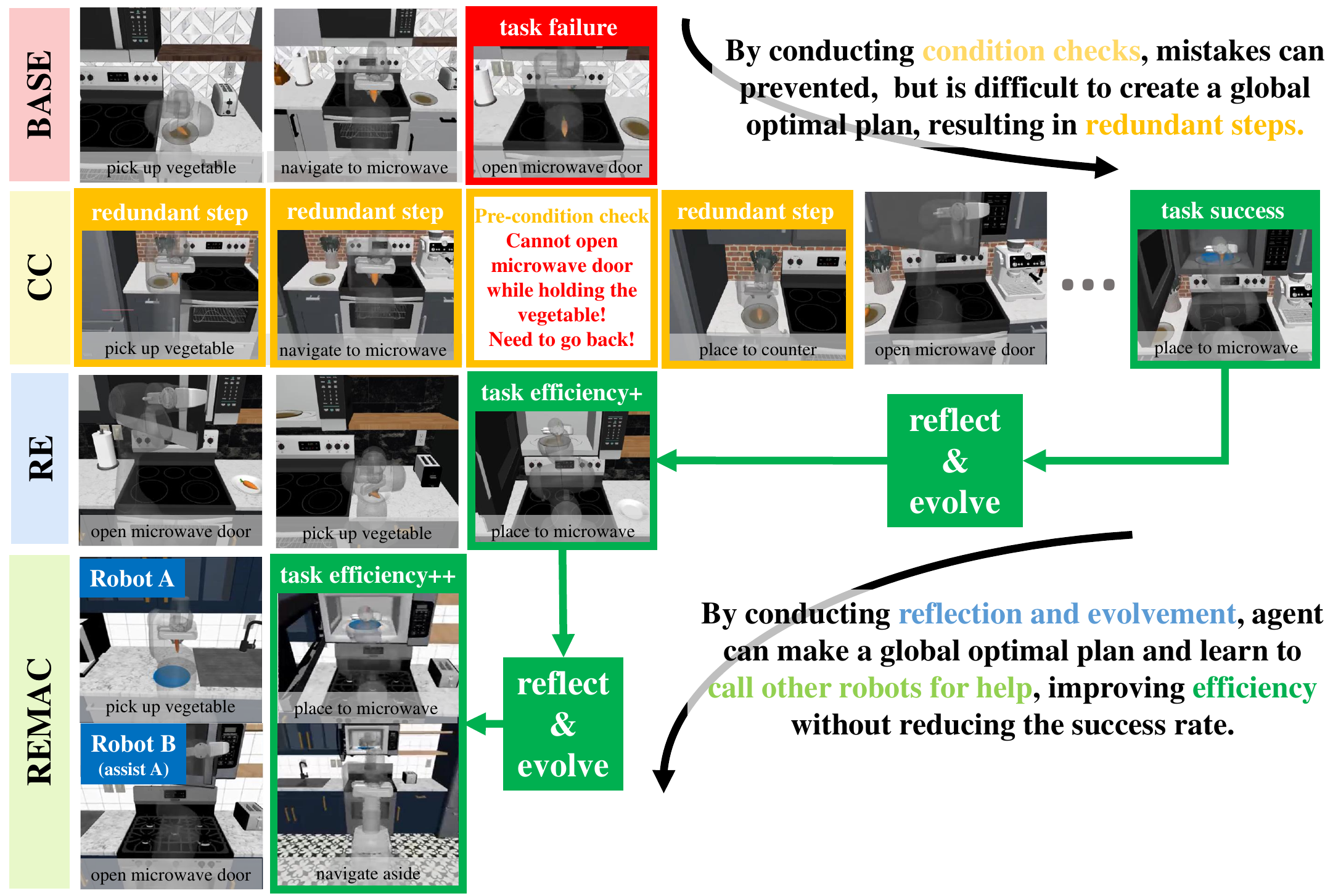}
  \caption{Typical pattern analysis, taking OpenMicrowave PnP as an example. \textbf{BASE setting:} The robot failed because it neglected the logical and spatial constraints of the environment, resulting in the carrot being dropped on the ground. \textbf{CC setting:} By incorperating condition checking mechenisms, the robot can successfully complete the task, but underwent a series of redundant steps before realizing it needs to put down the carrot first, resulting in a suboptimal plan. \textbf{RE setting:} With relfection and evolvement from past suboptimal plans, the robot can make a global optimal plan without redundant steps, thus completing the task with high efficiency. \textbf{REMAC setting:} Robot B is called independently to handle pre-conditions, assisting robot A to execute tasks with the main object, collaboratively completing the task with higher efficiency.
}
  \label{fig:failure}
\end{figure}

\begin{figure}
  \centering
  \includegraphics[width=0.99\linewidth]{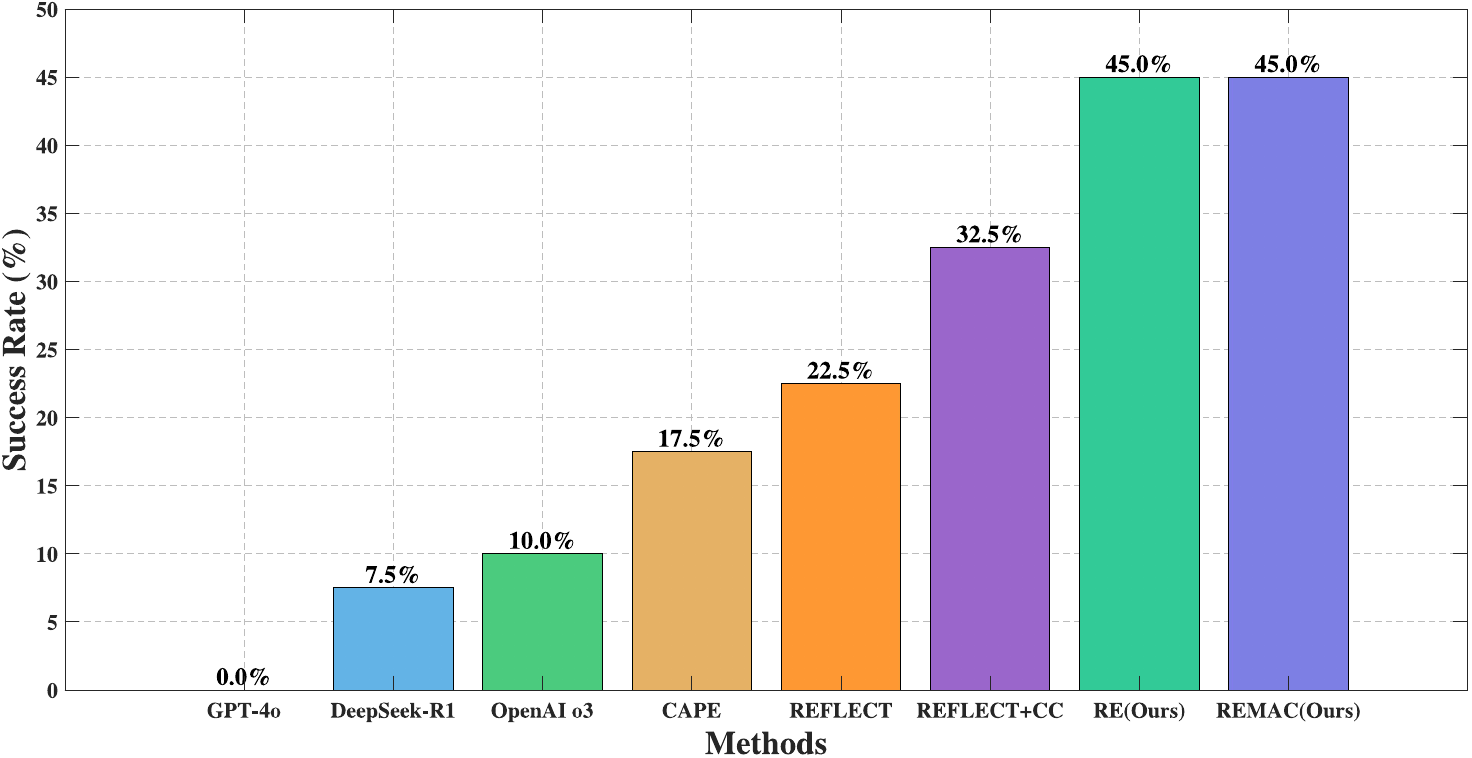}
  \caption{
  Comparison with baselines, where each data is the average task success rate of 20 experiments on 4 tasks. We use DeepSeek-R1 as the planning LLM in CAPE and REFLECT(+CC), and gpt-4o as the checking VLM in CAPE and REFLECT+CC. Settings of RE and REMAC are the same as previous.
  }
  \label{fig:baseline}
\end{figure}

Aligned with our framework's original design principles, transitioning from single-robot (RE) systems to multi-robot (\alias) coordination reduced average subtask duration by approximately 20\% through optimized task decomposition. While multi-robot systems demonstrated reduced robustness in certain scenarios - particularly due to failure propagation mechanisms where individual robot malfunctions could propagate through the system, leading to task failure - the coordinated approach maintained operational effectiveness with only marginal success rate reductions ($\leq 5.8$\%  absolute decrease). This performance profile confirms that multi-robot coordination successfully balances efficiency gains with acceptable robustness tradeoffs for executing extended-duration complex missions.

\begin{figure}
  \centering
  \includegraphics[width=1.0\linewidth]{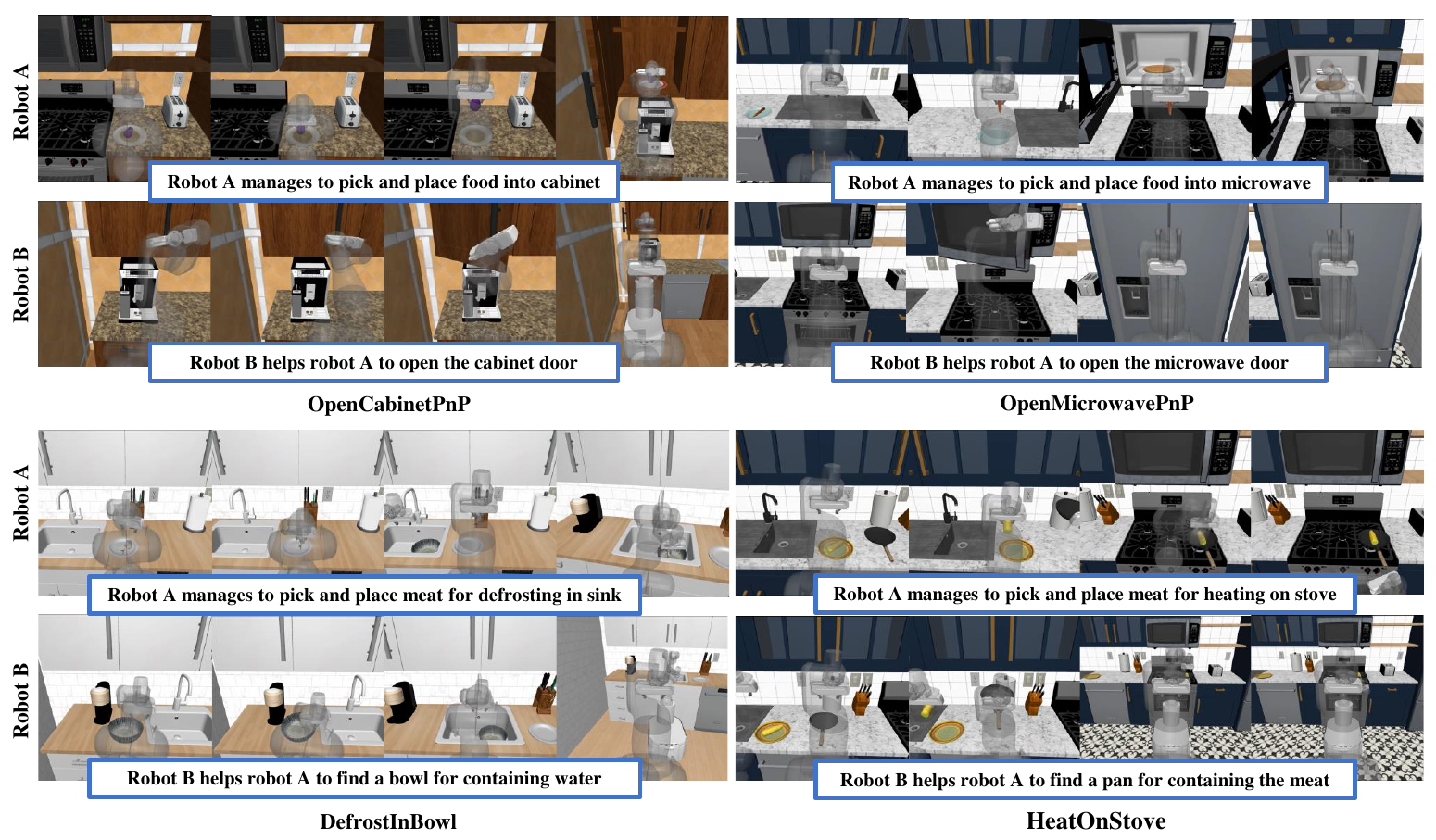}
  \caption{Trajectory visualization (key frames) from perspectives of two robots for four tasks (zoom in for a better view). Robot A serves as the main agent to execute main-object-centric plan, while robot B handles pre-conditions, assisting robot A to improve efficiency.
  \textbf{OpenCabinetPnP:} a robot calls another to open the cabinet.
  \textbf{OpenMicrowavePnP:} a robot requests another to open the microwave.
  \textbf{DefrostInBowl:} a robot directs another to place a bowl.
  \textbf{HeatOnStove:} a robot instructs another to place a pan.
  }
  \label{fig:two_robots}
\end{figure}

\begin{table}
\centering
\setlength{\tabcolsep}{12pt}
\caption{Reflect success rate for different reasoning models.}
\label{tab:model_compare}
\resizebox{\textwidth}{!}{%
\begin{tabular}{ccccccc}
\toprule
& \textbf{gpt-4o} & \textbf{QwQ} & \textbf{DeepSeek-R1} & \textbf{o3} & \textbf{Qwen3} & \textbf{Grok3} \\
\midrule
OpenCabinetPnP  & 15\% & 20\% & 40\% & 55\% & \textbf{80\%} & 70\% \\
OpenMicrowavePnP& 10\% & 25\% & 25\% & 55\% & 65\% & \textbf{85\%}  \\
DefrostInBowl   & 10\% & 40\% & 40\% & 60\% & 70\% & \textbf{95\%} \\
HeatOnStove     & 15\% & 20\% & 45\% & \textbf{80\%} & 50\% & \textbf{80\%} \\
\midrule         
Average         & 12.5\% & 26.25\% & 37.5\% & 62.5\% & 66.25\% & \textbf{82.5\%} \\
\bottomrule
\end{tabular}
}
\end{table}

Figure~\ref{fig:baseline} shows baseline comparisons. GPT-4o failed due to lack of reasoning ability, while DeepSeek-R1 and o3 perform better by capturing logical and spatial constraints. CAPE uses a pre-condition checking mechanism to avoid unfeasible actions and improves success rate. REFLECT achieves better initial planning with DeepSeek-R1, and CC improves execution success rates. Our RE and \alias framework further surpasses REFLECT + CC in task success rate.

Figure~\ref{fig:failure} shows an example of plan improvement from BASE to REMAC on the OpenMicrowavePnP task. Figure~\ref{fig:two_robots} demonstrates successful multi-robot collaboration for 4 tasks. 

\subsection{Realworld Experiments}


For the real-world experiments, we built a dual-arm mobile manipulation robot based on HEXMOVE and equipped it with two PiPER robotic arms. The robot uses a Femto Bolt RGB-D sensor as the head-mounted camera, together with two Gemini 336L RGB-D sensors mounted on the wrists to support manipulation execution. In addition, an Intel RealSense T265 tracking camera is employed to estimate the robot camera poses for reconstructing the deployment scene.

\begin{figure}
    \centering
    \includegraphics[width=0.5\linewidth]{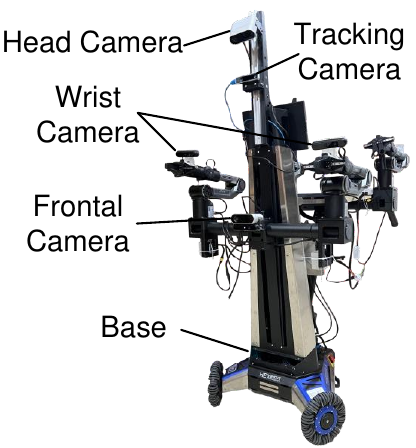}
    \caption{Real-world experiment platform.}
    \label{fig:real}
\end{figure}

Figure~\ref{fig:real} illustrates the real-world experimental platform. All real-world experiments were conducted on a single RTX 4060 GPU.

The real-world experiments reproduce the OpenMicrowavePnP task. The real-world experimental results are shown in Table~\ref{tab:phase_success}. BASE does not perform pre- and post-condition checking, and therefore fails to recognize that the plate must be put down before opening the door. CC introduces condition checking, but lacks the reflection-based iterative evolvement mechanism. Since VLMs can still make mistakes, CC cannot consistently identify the failure case where the door remains closed. In contrast, REMAC successfully incorporates such reflections into the database, enabling the system to avoid repeated failures and achieve a higher success rate. Please see our website link for a demonstration for CC (and also REMAC in the first iteration).

\begin{table}[t]
\centering
\caption{Success rate on real robot for each phase under different methods. We conduct 20 experiments and report average success rate. Phase1: open microwave. Phase2: pick and place vegetable into microwave. Phase3: add seasoning. Phase4: close microwave door. Time: average execution time of the final successful trajectories.}
\label{tab:phase_success}
\begin{tabular}{lccccc}
\toprule
\textbf{Method}
& \textbf{Phase1$\uparrow$}
& \textbf{Phase2$\uparrow$}
& \textbf{Phase3$\uparrow$}
& \textbf{Phase4$\uparrow$}
& \textbf{Time$\downarrow$} \\
\midrule
BASE  & 0\%  & 0\%  & 0\% & 0\%  &-- \\
CC    & 35\%  & 25\%  & 15\%  & 10\% &128.5s \\
REMAC & \textbf{55\%} & \textbf{40\%} & \textbf{35\%} & \textbf{25\%} & \textbf{57.2s} \\
\bottomrule
\end{tabular}
\end{table}

\subsection{Reflective Abilities Among Reasoning Models}

Moreover, we compare the reflection ability for different reason models, including DeepSeek-R1~\cite{guo2025deepseek}, o3~\cite{openai2025o3}, QwQ~\cite{qwen2, qwen2.5}, Qwen3~\cite{yang2025qwen3}, and Grok3~\cite{xai2025grok3}. 

Given previous iterations and reflections, models must reflect on non-optimal plans and output refined plans. Models with reasoning capabilities significantly outperform those without. Reflection success rate measures a model's ability to generate optimal plans from suboptimal ones (Table~\ref{tab:model_compare}). All tasks used 20 randomized initializations across models. Grok3 outperforms Qwen3, o3, DeepSeek-R1, and QwQ, showing superior reflection ability.

Notably, large language models without reasoning capabilities are highly sensitive to task descriptions; minor word order changes lead to different planning outcomes. These models poorly attend to previous reflections, hindering effective iterative evolvement.

\section{Conclusions}

We propose \alias, a task planning framework established within the RoboCasa large-scale simulation environment, integrating self-reflection and self-evolvement methodologies. This framework addresses the challenging setting of long-horizon multi-robot task planning by leveraging information from pre-condition checks to iteratively refine the initial plan and employing post-condition checks to guide retry attempts. As a result, it substantially enhances both success rates and operational efficiency in an self-evolving manner. Our work resolves two critical limitations in existing approaches: \textbf{1)} The tendency of VLM to neglect logical and spatial constraints within the environment during task planning under ambiguous instructions and \textbf{2)} The inability of prevailing multi-robot task planning methods to handle tasks beyond short-duration horizons effectively. Future research directions may include adopting a longer pre-condition checking horizon or exploring collaboration between heterogeneous robots.

\bibliographystyle{elsarticle-num}
\bibliography{cas-refs}

\end{document}